\pdfoutput=1
\documentclass[11pt]{article}

\usepackage{authblk}

\usepackage{ACL2023}
\usepackage[ruled,vlined]{algorithm2e}
\usepackage{times}
\usepackage{latexsym}
\usepackage{graphicx}
\usepackage[T1]{fontenc}
\usepackage{subcaption}
\usepackage{booktabs}
\usepackage{amsmath}
\usepackage[utf8]{inputenc}

\usepackage{microtype}

\usepackage{inconsolata}

\usepackage{amsfonts}
\usepackage{etoolbox}
\title{CAME: Confidence-guided Adaptive Memory Efficient Optimization}

\author[1*]{\textbf{Yang Luo}}
\author[2]{\textbf{Xiaozhe Ren}}
\author[1]{\textbf{Zangwei Zheng}}
\author[1]{\textbf{Zhuo Jiang}}
\author[2]{\textbf{Xin Jiang}}
\author[1]{\textbf{Yang You}}
\affil[1]{School of Computing, National University of Singapore}
\affil[2]{Noah’s Ark Lab, Huawei}
\affil[ ]{\texttt{\{yangluo,zangwei,jiangz,youy\}@comp.nus.edu.sg}}
\affil[ ]{\texttt{\{renxiaozhe,jiang.xin\}@huawei.com}}

\begin{document}
\maketitle
\begingroup\def\thefootnote{*}\footnotetext{Work was done when Yang Luo was an intern at Huawei Noah’s Ark Lab.}\endgroup
\begin{abstract}
Adaptive gradient methods, such as Adam and LAMB, have demonstrated excellent performance in the training of large language models. Nevertheless, the need for adaptivity requires maintaining second-moment estimates of the per-parameter gradients, which entails a high cost of extra memory overheads. To solve this problem, several memory-efficient optimizers (e.g., Adafactor) have been proposed to obtain a drastic reduction in auxiliary memory usage, but with a performance penalty. In this paper, we first study a confidence-guided strategy to reduce the instability of existing memory efficient optimizers. Based on this strategy, we propose CAME to simultaneously achieve two goals: fast convergence as in traditional adaptive methods, and low memory usage as in memory-efficient methods. Extensive experiments demonstrate the training stability and superior performance of CAME across various NLP tasks such as BERT and GPT-2 training. Notably, for BERT pre-training on the large batch size of 32,768, our proposed optimizer attains faster convergence and higher accuracy compared with the Adam optimizer. 
The implementation of CAME is publicly available\footnote{\url{https://github.com/yangluo7/CAME}}.
\end{abstract}

\section{Introduction}
Robust training of large language models (LLMs) often relies on adaptive gradient-based optimization methods \cite{pmlr-v162-li22b, DBLP:journals/corr/KingmaB14, zhuang2020adabelief}. Through the use of cumulative second-order statistics, these methods adapt the per-parameter learning rate and demonstrate superior convergence speed during the training process of LLMs. However, the remarkable performance of adaptive methods incurs an extra cost of memory usage indeed. For example, Adam requires to preserve the first moment estimate and second raw moment estimate of each gradient in order to tune the learning rate for each parameter, which inevitably triples the memory usage concerning the optimizer states. Besides, with the growing size of the model, LLMs are becoming increasingly expensive in terms of memory, and the limitation of memory is gradually emerging as a main bottleneck for training LLMs. 

% Over the past few years, large language models (LLMs) trained with adaptive gradient optimization methods show strong abilities of multi-task and few-shot learning, and achieve remarkable performances on various natural language processing tasks. However, Due to the large number of parameters, LLMs are typically expensive in terms of memory and the limitation of memory is gradually emerging and becomes a main bottleneck for training LLMs. 

 \begin{figure}[t]
    \centering
    \includegraphics[width=0.5\textwidth]{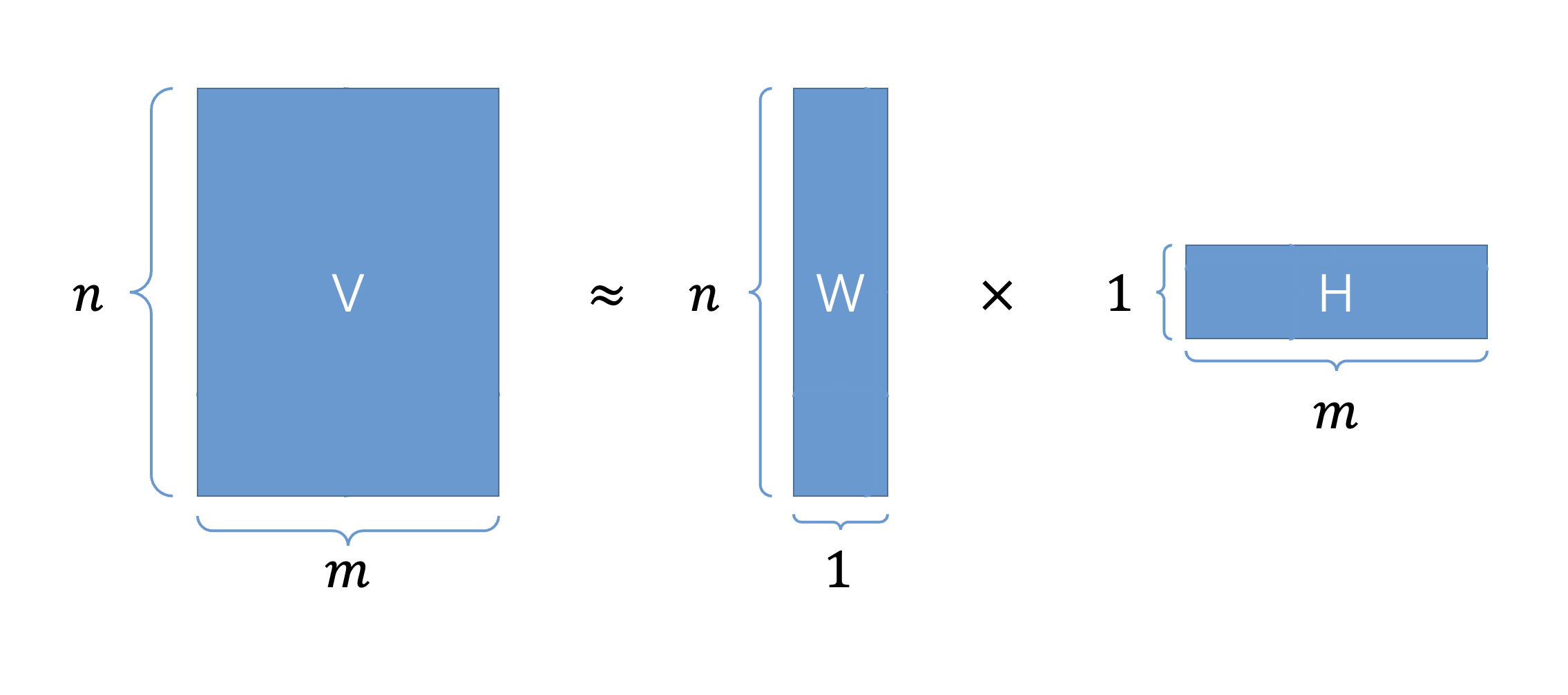}
    \caption{Visualization of Non-negative Matrix Factorization (NMF). Generally, NMF reduces the memory requirements from $O(nm)$ to $O(n + m)$. In this paper, we focus on the special case of rank-1 factors. 
    % Trajectories of SGD, Adam and AdaBelief. AdaBelief reaches optimal point (marked as
% orange cross in 2D plots) the fastest in all cases. We refer readers to video examples.
}
    \label{fig:nmf}
\end{figure}

Many existing memory-efficient optimizers attempt to store second-order statistics with sublinear memory requirement while retaining the exceptional convergence property of adaptivity \cite{Shazeer2018AdafactorAL, 10.5555/3454287.3455161}. Adafactor optimizer achieves remarkable memory cost reduction by applying the non-negative matrix factorization algorithm \cite{nmf} to factorize the accumulator matrix for squared gradients into two rank-1 factors as shown in Figure \ref{fig:nmf}, where the memory requirement for the original matrix $V$ decreases from $O(nm)$ to $O(n+m)$. Whereas, it is observed that Adafactor suffers a performance degradation in the training of large language models universally compared with conventional adaptive gradient-based optimization methods.  The reason for this phenomenon is Adafactor inevitably introduces some errors that cause instability in training deep networks due to the operation of non-negative matrix factorization.
 
 In addition, in the case of large-batch training that aims to accelerate the training of deep neural networks, the memory consumption of each machine (GPU/TPU) is much higher than general batch size training, which further imposes a grave constraint on the performance of the trained model. In comparison to standard training tasks, large-batch training presents more challenges for optimizers. Empirically, 
 % the test accuracy of the converged solution generally becomes significantly lower than the baseline when mini-batch size increases after a certain point (e.g. 1024).
 when the mini-batch size increases after a certain point (e.g. 1024), the test accuracy of the converged solution decreases significantly compared with the baseline \cite{https://doi.org/10.48550/arxiv.2111.00856}.
 To our knowledge, there is currently no work related to memory-efficient optimizers for large-batch training.
 
Motivated by these challenges, we firstly study a confidence-guided strategy catered to alleviate the instability of Adafactor by calculating the confidence of the generated update at each training step. On the basis of the adaptation strategy, we propose a novel CAME optimizer that saves nearly the same memory footprint as existing memory-efficient optimizers while attaining faster convergence and superior generalization performance. To further assess the scalability of our proposed algorithm, we consider an additional challenging experiment - performing large-batch training on BERT using CAME optimizer.

Contributions of our paper can be summarized in the following:
\begin{itemize}
    \item Inspired by training instability of Adafactor, we explore a confidence-guided strategy centered on the existing error in the raw updates of Adafactor for parameters of large language models.
    \item In light of the dedicated strategy, we propose a novel optimization algorithm, CAME, for achieving faster convergence and less performance degradation catered at memory-efficient optimization. We further investigate the effect of the proposed memory-efficient optimization algorithm in large-batch training settings.
    \item We demonstrate the powerful performance of CAME with extensive NLP experiments: CAME shows faster convergence and better generalization capability than Adam in BERT pre-training task with two different batch sizes (32k and 8k); in the training of GPT-2 model and T5 model, CAME achieves fast convergence speed as Adam without degrading of performance. Notably, in the large-batch training of the BERT model, CAME obtains comparable validation accuracy with LAMB using around 15\% less memory usage.
\end{itemize}

\section{Related Work}
% \textbf{Adaptive Gradient Algorithms} Adaptive gradient-based algorithms generally forms essential parts of training modern deep neural networks in a wide range of tasks. 

\textbf{Memory Efficient Adaptive Optimization} 
% Adaptive gradient-based algorithms generally forms essential parts of training modern deep neural networks in a wide range of tasks. However, most existing optimizers (e.g.,  Adam \cite{DBLP:journals/corr/KingmaB14}, Adabelief \cite{zhuang2020adabelief}, LAMB \cite{lamb}) will triple the required memory footprint because of keeping first and second estimates, which serves a main barrier for large-scale language models training. 
Memory efficient optimizers maintain the benefits of standard per-parameter adaptivity while significantly reducing memory footprint.
Adafactor \cite{Shazeer2018AdafactorAL} proposes to reconstruct a low-rank approximation of the exponentially smoothed accumulator at each training step that is optimal with respect to the generalized Kullback-Leibler divergence. 
% In the case of an $n \times m$ matrix, Adafactor reduces the memory requirement from $O(mn)$ to $O(m+n)$. 
SM3 \cite{10.5555/3454287.3455161} divides the elements in the second-order gradient matrix into sets by the observed similarity of the elements, and each item in the generated approximation matrix is the minimum of the maximum value of each set in which it is located. The methods mentioned above behave poorly in the training of large language models and converge slowly, which raises a significant challenge for memory-efficient optimization methods.

% Another closely related method is the Shampoo [10] algorithm for optimization over tensor structures. Seemingly, the goal of Shampoo is very different from ours: going beyond entrywise learning rates and employing full-matrix regularization in a computationally efficient way. Nonetheless, Shampoo can also be seen as a method to substantially reduce the memory footprint of full-matrix preconditioned algorithms (specifically, full-matrix Adagrad). In a sense, our algorithms are analogous to a diagonalized version of the Shampoo algorithm. Yet another recent adaptive optimization method is the GGT algorithm [2]. Similarly to Shampoo, the goal of the latter is to reduce the computation cost of full-matrix preconditioning in order to make it practical in large scale settings. However, GGT stores multiple copies of the gradient over the course of its execution, and as a result, its space requirements restricts it from being applied at large scale.

\textbf{Large Batch Training} A large-batch training scheme is preferred in distributed machine learning because of its ability to increase parallelism by enhancing large-scale cluster utilization.
% an important direction for distributed machine learning, which can improve the utilization of large-scale clusters and accelerate the training process. 
It has seen growing interest in recent years in large-batch training \cite{looksam, 1-bit_lamb, Huo_Gu_Huang_2021}. 
 In particular, a layer-wise adaptive learning rate algorithm LARS \cite{lars} is proposed  to scale the batch size to 32k for ResNet-50. Based on LARS, LAMB optimizer \cite{lamb} can finish the BERT training in 76 minutes through TPU v3 Pod. Despite the success of these approaches for BERT models, the much larger batch size highly boosts the GPU usage which is prohibitively expensive and inaccessible to most researchers. 

%  However, there is no exploration of memory efficient optimizers for large-batch training.
Moreover, training with a large batch size incurs additional challenges \cite{Train_longer, https://doi.org/10.48550/arxiv.1609.04836}.
Large-batch training is prone to converge to sharp local minima, since the number of interactions will decrease when the batch size is increased if the number of epochs is fixed, which causes a wide gap in generalization of the model\cite{https://doi.org/10.48550/arxiv.1609.04836}. Traditional methods seek to narrow the generalization gap by carefully tuning hyperparameters, such as learning rate, momentum, and label smoothing, to narrow the generalization gap \cite{DBLP:journals/corr/GoyalDGNWKTJH17, https://doi.org/10.48550/arxiv.1811.03600, https://doi.org/10.48550/arxiv.1709.05011}.
Yet there have been few attempts to reduce memory usage in large-batch training, and the underlying challenge remains unclear.
% However, these heuristic approaches cannot be regarded as a principle solution for large-batch training \cite{https://doi.org/10.48550/arxiv.1811.03600}.

% Recently, to avoid these hand-tuned methods, adaptive learning rate on large-batch training has gained enormous attention from researchers \cite{NEURIPS2018_90365351, https://doi.org/10.48550/arxiv.1912.03194}. 
% % Many recent works attempt to use adaptive learning rate to scale the batch size for BERT on ImageNet [1, 5, 8, 18, 22, 32, 34, 39, 42, 47, 48].
% In particular, layer-wise adaptive learning rate algorithm LARS \cite{lars} is proposed  to scale the batch size to 32k for ResNet-50. Based on LARS optimizer, LAMB optimizer \cite{lamb} can finish the BERT training in 76 minutes through TPU v3 Pod \cite{lamb}. However, there is no exploration of memory efficient optimizers for large-batch training.
% Moreover, \cite{lamb} proposes the LAMB optimizer to scale up the batch size when training ResNet-50, resulting in a 2.2 minutes training time.

\section{Method}
\label{sec:eu}
In this section, we firstly provide a brief description of the Adafactor optimizer and discuss the errors contained in the update of Adafactor (erroneous update). We further study a confidence-guided strategy and introduce the proposed CAME in detail in light of the strategy.

\subsection{An overview of Adafactor}

% We firstly review the Adafactor optimizer in this section for reference. 
The $\mathcal{L}(\theta) \in \mathbb{R}$ represents the loss function that we plan to minimize, where $\theta \in \mathbb{R}^{n \times m}$ is the parameter of the model. $g_{t}$ is the gradient at step $t$, $\eta$ is the learning rate, $r_{t}$  and $c_{t}$ are the exponential moving average of two low-rank factors for the second moments of the gradient. $\epsilon_{1}$ is a small regularization constants and $u_{t}$ is the current approximate update. 

% \SetKwBlock{Storing}{Backbone learning and HG Bank Filling}{end}
\SetKwBlock{Training}{Adversarial HARDer-Net Learning}{end}
\SetKwBlock{DQN}{Hardness-Guided Learning}{end}
\begin{algorithm*}[t]
%\scriptsize
	\SetAlgoLined
%	\BlankLine
	\KwIn{Initial parameters $\theta_0$, learning rate $\eta$, momentum of update $m_0$, $r_0$, $c_0$, step $t$, regularization constant $\epsilon_1$, exponential moving average parameters {$\beta_1, \beta_2$}, clipping threshold $d$}
	\While{$\theta_t$ not converge}{
% 		\Storing{
        Compute $g_t$ = $\nabla f(\theta_{t-1})$ \\
        $r_t = \beta_2 r_{t-1} + (1-\beta_2) (g_t^{2} + \epsilon_1 1_n 1^T_m)1_m$\\
        $c_t = \beta_2 c_{t-1} + (1-\beta_2) 1^T_n (g_t^{2} + \epsilon_1 1_n 1^T_m)$\\
        $v_t = r_t c_t / 1^T_n r_t$ \\
        $u_t = g_t / \sqrt{v_t}$ \\
        $\hat{u}_t = u_t/$max$(1, RMS(u_t)/d)$ \\
        $m_t = \beta_1 m_{t-1} + (1 - \beta_1) \hat{u}_t$ \\
        $\theta_t = \theta_{t-1} - \eta m_t$ \\
% 		}
		}
	
	\caption{Adafactor Optimizer}
	\label{alg:adafactor}
\end{algorithm*}

In the training of large language models, Adafactor is required to apply momentum to ensure the convergence \cite{chowdhery2022palm}, and the corresponding pseudocode is illustrated in Algorithm \ref{alg:adafactor}. 
The problem setting is as follows.
Assume that we aim to minimize the expected value of an objective function $f(\theta)$. At each training step, we receive the loss derived from a mini-batch of data, and calculate the gradient $g_{t}$ of the function based on the previous parameters. Subsequently, we update the exponential running averages of two factors for second moments of the gradient $r_t$ and $c_t$, compute approximations for the second moments of the gradient $v_t$, and adjust the generated update ($u_t$) when $RMS(u_t)$ surpasses a specific threshold value $d$ as in:
\begin{equation}
    \hat{u}_t = \frac{u_t}{\text{max}(1, RMS(u_t)/d)}
\end{equation}
where $RMS(u_t)$ refers to the root-mean-square calculation of the components of $u_t$.
Finally, the first moment of the adjusted update $m_t$ is utilized to update the parameter, resulting in a new iteration $\theta_t$. The optimization continues until the parameters converge and returns the final iteration $\theta_T$ as our approximate solution.

Adafactor derives an effective solution for non-negative matrix factorization in the special case of rank-1 factors, which obtains the minimal Kullback–Leibler divergence \cite{lee99} between the matrix $V$ and the approximated matrix $WH$. The formulation of the solution is as follows, in which $1_m =(1,...,1) \in \mathbb{R}^m$ represents a column vector of $m$ ones:
\begin{equation}
    W = V1_m, \quad H = \frac{1_n^{T}V}{1_n^{T}V1_m}.
\end{equation}

\begin{figure}[t]
    \centering
    \includegraphics[width=0.5\textwidth]{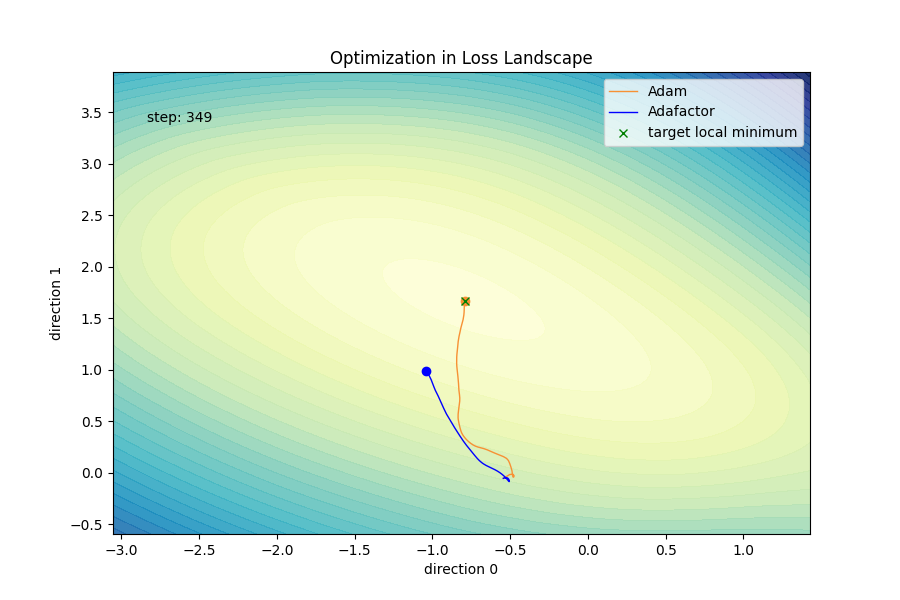}
    \caption{Loss landscape visualization for erroneous update of Adafactor in 1-layer multilayer perceptron (MLP) \cite{mlp} with same training steps. Adafactor deviates from the training curve of Adam.}
    \label{fig:adafa_eu}
\end{figure}
It should be noted that Adafactor stores only the moving averages of these factors rather than the entire matrix $V$, yielding considerable memory savings and requiring memory usage proportional to $O(n + m)$ instead of $O(nm)$.

\subsection{Erroneous Update}
The non-negative matrix factorization operation in Adafactor will inevitably incur erroneous update in the training of deep neural networks. As shown in Figure \ref{fig:adafa_eu}, Adafactor always converge slower than Adam due to the existing error in calculated updates, which further limits the application scenarios of memory-efficient optimizers. 

\begin{figure}[hbt!]
    \centering
    \begin{subfigure}[b]{0.5\textwidth}
        \centering
        \includegraphics[height=1.5in]{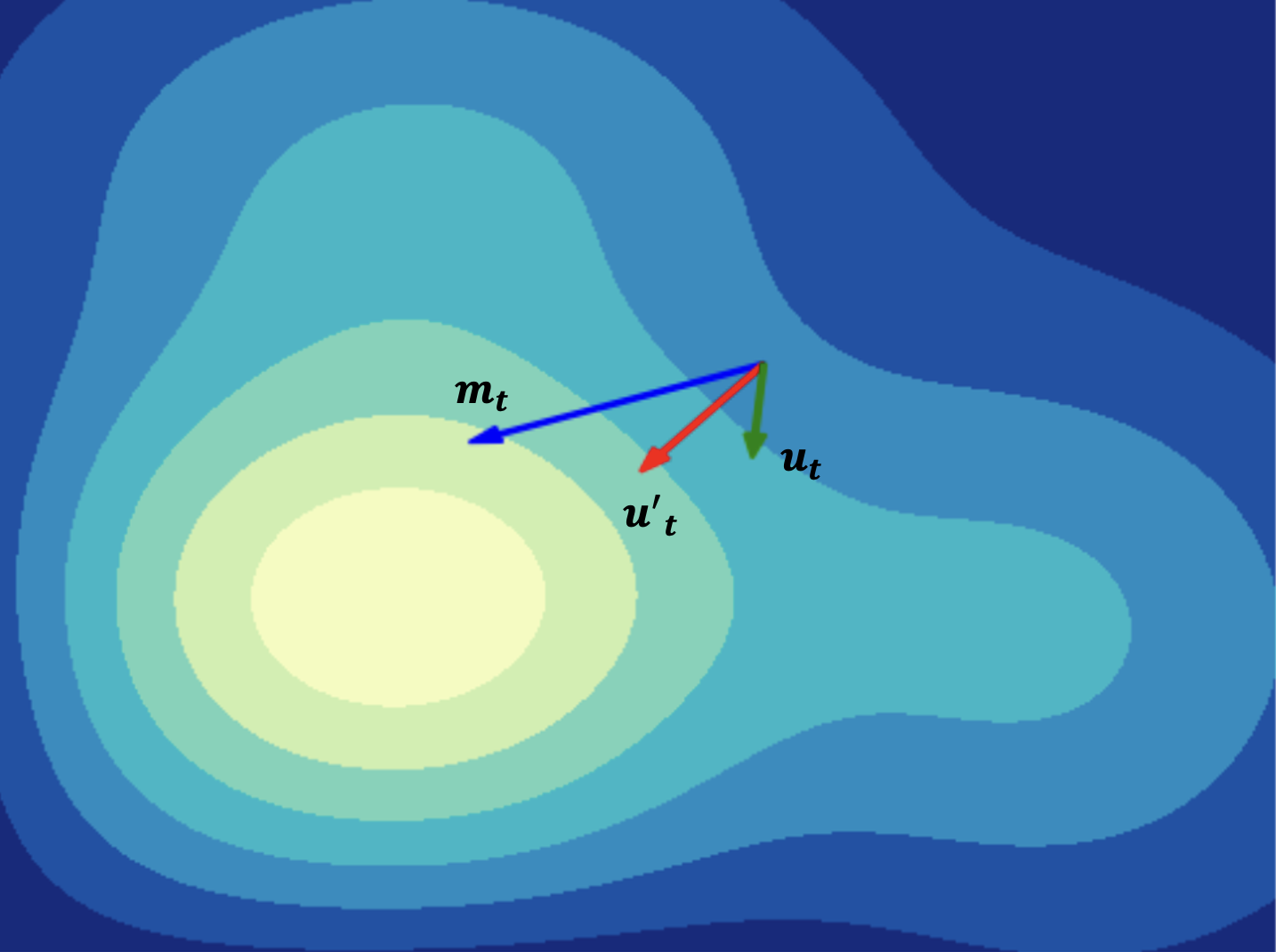}
        \caption{scenario 1}
        \label{s1}
    \end{subfigure}%
    
    \begin{subfigure}[b]{0.5\textwidth}
        \centering
        \includegraphics[height=1.5in]{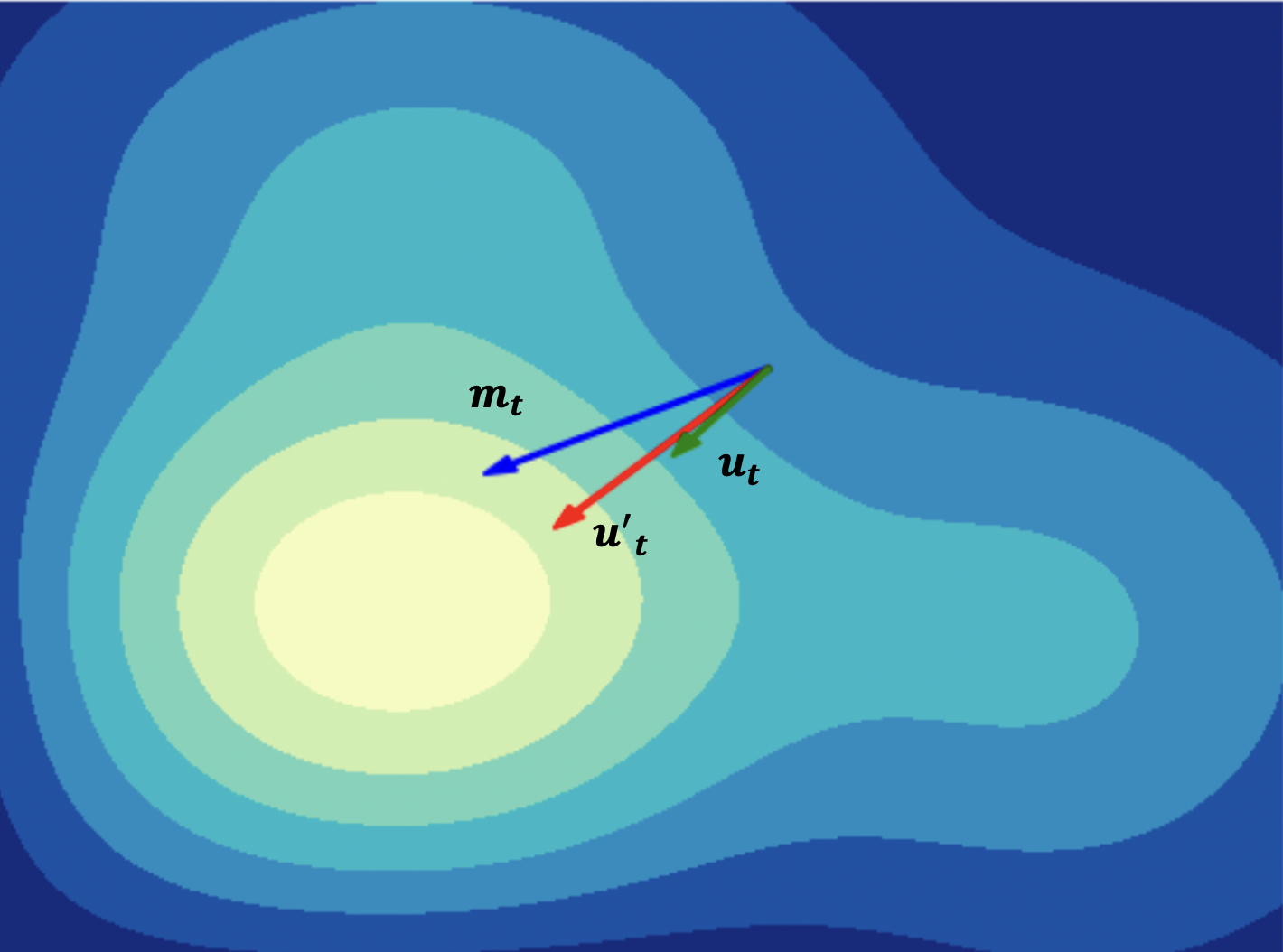}
        \caption{scenario 2}
        \label{s2}
    \end{subfigure}
    \caption{Visualization of two scenarios where Adafactor updates have different stability. }
    \label{fig:updates}
\end{figure}

As shown in Figure \ref{fig:updates}, two scenarios demonstrate how two types of erroneous updates are supposed to be handled in the ideal case.  
In Figure \ref{fig:updates}(a), the difference between the momentum of updates $m_t$ and the current update $u_t$ is large, illustrating that the historical experience for the update of original Adafactor contains high level of errors that will inevitably influence the stability of the training process. If we utilize the raw $m_t$ to take an optimization step, the direction of optimization will deviate increasingly from the desired direction, which is reflected by the slow convergence and performance degradation of existing memory-efficient optimizers. By contrast, when the difference between $m_t$ and $u_t$ is small as shown in Figure \ref{fig:updates}(b), the momentum $m_t$ is stable with limited errors and high confidence therefore a large optimization step is required with the updating direction close to $m_t$.

Inspired by the erroneous update that is universal in existing memory-efficient optimizers, we firstly consider an efficient approach to decrease the side effect caused by insecure updating. Given $m_t$ and $u_t$, we take the residual between them as the instability in the preserved momentum and set generated instability as the denominator of original $m_t$ to more adaptively take an update step. Following is the formulation of the adjusted update $u'$, where $\epsilon$ is the regularization constant:
\begin{equation}
    u'_t = \frac{m_t}{\sqrt{(m_t-u_t)^2 + \epsilon}}
\end{equation}

% Under ideal conditions, if the involved error of the momentum $m_t$ is large, we aim to update parameters of the deep neural network less. On the contrary, if the error is small, as shown in Fig. \ref{fig:updates}(b), which demonstrates the update is stable with limited error, the optimal update obligate to be large.

Extending further on the plain method, we propose a confidence-guided strategy that enables self-adjusted updates by taking the confidence of the raw update of Adafactor into consideration. The intuition behind the proposed strategy is to calculate the residual between the exponential moving average (EMA) of the update and the current update, which represents the deviation of the approximated update. The larger the deviation of the EMA value from the current generated update, the wider the error EMA of update contains, resulting in a lower level of confidence in the EMA of update. Obviously, we expect the optimizer to take a small update when it incorporates huge error (a large residual from the present update), while updating parameters more when the optimization process is stable (involved error of EMA is limited). 

Specifically,
% by denoting the EMA of $(u_t - m_t)^2$ as $\textit{Conf}_t$, respectively.
the EMA of update $m_t$ is directly used to take an update step in Adafactor, while in our proposed strategy, $m_t$ is divided by $\sqrt{\textit{U}_t}$, where $\textit{U}_t$ is the calculated instability matrix. Therefore, $\frac{1}{\sqrt{\textit{U}_t}}$ is the confidence in the observation: viewing $m_t$ as the prediction of the update, if $m_t$ deviates greatly from $u_t$ ($\textit{U}_t$ is large), which indicates a weak confidence in $m_t$, the optimizer performs a small optimization step; if $u_t$ closely matches $m_t$, we have solid confidence in $m_t$, and correspondingly take a large optimization step.

\subsection{CAME Algorithm}
\begin{algorithm*}[t]
%\scriptsize
	\SetAlgoLined
%	\BlankLine
	\KwIn{Initial parameters $\theta_0$, learning rate $\eta$, momentum of update $m_0 = 0$, $r_0 = 0, c_0 =0$, step $t$ = 0, regularization constants $\epsilon_1, \epsilon_2$, exponential moving average parameters {$\beta_1, \beta_2, \beta_3$}, clipping threshold $d$}
	\While{$\theta_t$ not converge}{
% 		\Storing{
        Compute $g_t$ = $\nabla f(\theta_{t-1})$ \\
        $r_t = \beta_2 r_{t-1} + (1-\beta_2) (g_t^{2} + \epsilon_1 1_n 1^T_m)1_m$\\
        $c_t = \beta_2 c_{t-1} + (1-\beta_2) 1^T_n (g_t^{2} + \epsilon_1 1_n 1^T_m)$\\
        $v_t = r_t c_t / 1^T_n r_t$ \\
        $u_t = g_t / \sqrt{v_t}$ \\
        $\hat{u}_t = u_t/$max$(1, RMS(u_t)/d)$ \\
        $m_t = \beta_1 m_{t-1} + (1 - \beta_1) \hat{u}_t$ 
        \textcolor{blue}{\\
        $\textit{U}_t = (\hat{u}_t - m_t)^{2}$ \\
        $R_t = \beta_3 R_{t-1} + (1-\beta_3) (\textit{U}_t + \epsilon_2 1_n 1^T_m)1_m$\\
        $C_t = \beta_3 C_{t-1} + (1-\beta_3) 1^T_n (\textit{U}_t + \epsilon_2 1_n 1^T_m)$\\
        $S_t = R_t C_t / 1^T_n R_t$ \\
        }
        $\theta_t = \theta_{t-1} -  \frac{\eta}{
        \textcolor{blue}{\sqrt{S_t}}} m_t$ \\
% 		}
		}
	
	\caption{CAME Optimizer}
	\label{alg:adares}
\end{algorithm*}

Based on the proposed confidence-guided strategy, we develop a brand-new variant of memory-efficient optimization methods with faster convergence. Our proposed CAME optimization method successfully obtains the same rate of convergence as prevailing  first-order optimization algorithms (e.g., Adam) and with almost equal memory cost to available memory-efficient optimizers (e.g., Adafactor). The pseudocode of CAME algorithm is specified in Algorithm \ref{alg:adares}.

By calculating $\textit{U}_t$ at each training step, 
% \textit{non-negative matrix factorization}, a popular technique applied in practice that factorizes a matrix $V$ into two matrices $W$ and $H$ with a smalle memory footprint, r
we employ non-negative matrix factorization on the instability matrix $\textit{U}_t$ following \cite{Shazeer2018AdafactorAL} where the generalized Kullback-Leibler divergence between $V$ and $WH$ is minimal. With $\textit{U}_t$ factorized into $R_t$ and $C_t$, it is sufficient to store only the moving averages of these factors rather than the full matrix $\textit{U}_t$, thus saving considerable memory footprint.

\begin{figure}[t]
    \centering
    \includegraphics[width=0.5\textwidth]{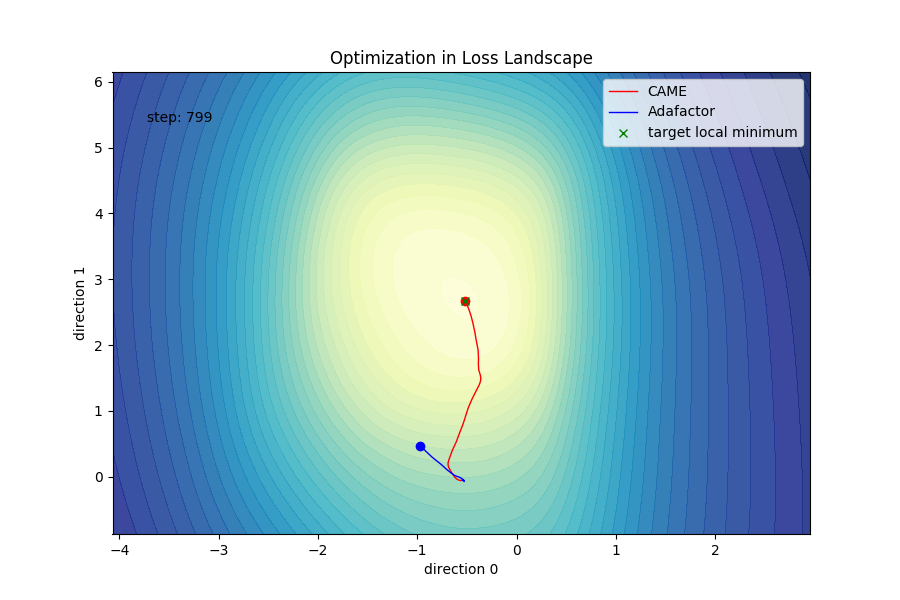}
    \caption{Loss trajectories of Adafactor and CAME. CAME reaches the target local minimum (marked as
green cross in 2D plots) much faster than Adafactor.}
    \label{fig:adafa_i}
\end{figure}

We simply validate intuitions and the corresponding example is shown in Figure \ref{fig:adafa_i}, in which the proposed CAME reaches the optimal point much faster than Adafactor. Learning rate is $10^{-3}$ for all optimizers. In the example, we set the parameters of CAME to be the same as the default in Adafactor, $\beta_1=0.9, \beta_2=0.999$ and set extra $\beta_3=0.9999$ for CAME.

\section{Experiments}

In this section, we present extensive comparisons with existing optimizers on training tasks of three important large language models: BERT \cite{devlin-etal-2019-bert}, GPT-2 \cite{noauthororeditor} and T5 \cite{T5}. 

\subsection{Setup}

\textbf{Dataset} We perform experiments on the BookCorpus \cite{noauthororeditor} and English Wikipedia with 800M and 2.5B words respectively. Furthermore, we focus on the GLUE benchmark \cite{peters2018contextualized}, SQuAD v1.1 dataset \cite{rajpurkar-etal-2016-squad} and SQuAD v2.0 dataset \cite{squadv2.0} to demonstrate the performance of pre-trained BERT models with CAME optimizer. 

\textbf{Model} We evaluate the efficiency of our proposed CAME on three trending large language models: BERT, GPT-2 and T5. We further test the performance of CAME for large-batch training with BERT-Large.

\textbf{Compared methods} 
% \begin{itemize}
%     \item Adafactor
%     \item Adam
%     \item LAMB 
% \end{itemize}
% \textbf{Baseline}\quad
The main baselines comprise two widely-used optimizers: classic optimizer Adam and memory-efficient optimizer Adafactor. With regard to large-batch training, LAMB optimizer is additionally considered when setting baselines.

\textbf{Implementation Detail} We implement our optimization algorithm in Pytorch \cite{pytorch}. The parameters $\beta_1$ and $\beta_2$ in Algorithm \ref{alg:adares} are set as 0.9 and 0.999 respectively, and we search for optimal $\beta_3$ among \{0.9, 0.99, 0.999, 0.9999, 0.99999\}. We use 8 Tesla V-100 GPUs and set $\epsilon_1$, $\epsilon_2$ as $10^{-30}$, $10^{-16}$ in all experiments with gradient accumulation and model parallelism. Besids, we set $\eta$ as $2 \times 10^{-4}$, $6 \times 10^{-4}$, $3 \times 10^{-4}$ for BERT-Large (32K), GPT-2, T5 training and apply learning rate warmup scheduling \cite{warmup} to avoid divergence due to the large learning rate, by starting with a smaller learning rate $\eta$ and gradually increasing to the large learning rate $\eta$. To make sure we are comparing with solid baselines, we use grid search to tune the
hyperparameters for Adafactor, Adam and LAMB. We further improve the performance of large-batch training by applying Mixup \cite{mixup} to scale the batch size up to 32,768.

\subsection{BERT Training}
\label{sec:bert}

\begin{figure}[t]
    \centering
    \includegraphics[width=0.5\textwidth]{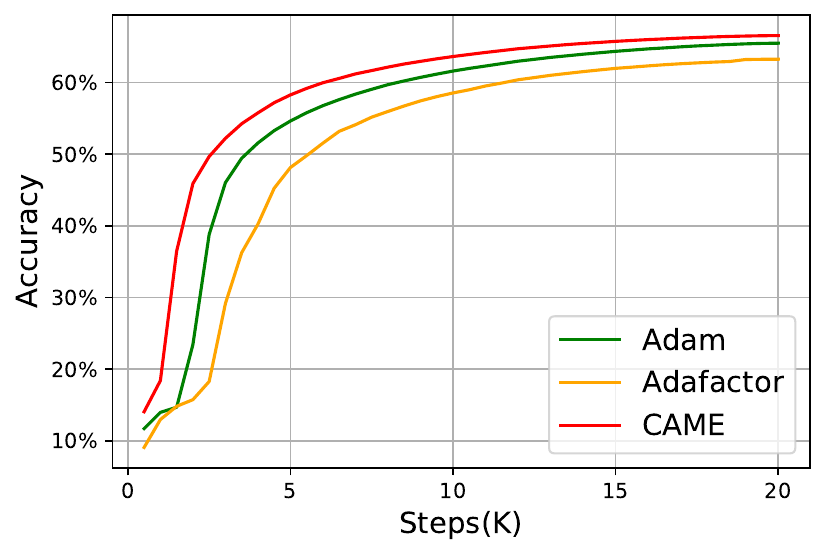}
    \caption{Masked LM test accuracy of BERT-Large model trained on Wikipedia dataset with 8k batch size.}
    \label{fig:r1}
\end{figure}

\begin{figure}[t]
    \centering
    \includegraphics[width=0.5\textwidth]{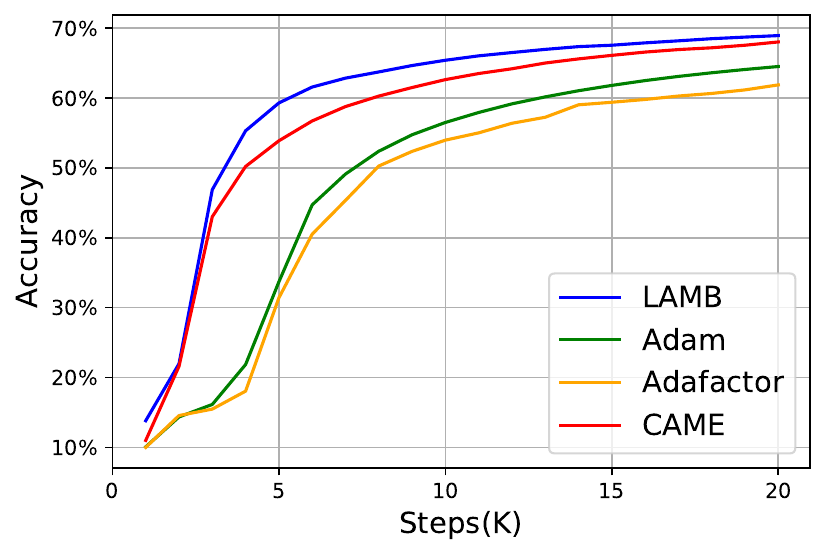}
    \caption{Masked LM test accuracy of BERT-Large model trained on Wikipedia dataset with 32k batch size. CAME achieves comparable accuracy with Adafactor using around only half of required training steps (10k).
    % Trajectories of SGD, Adam and AdaBelief. AdaBelief reaches optimal point (marked as
% orange cross in 2D plots) the fastest in all cases. We refer readers to video examples.
}
    \label{fig:r2}
\end{figure}

We firstly present empirical results in the training task of BERT model to evaluate the performance of our proposed CAME optimizer, focusing on its larger variant, BERT-Large, which has 340M parameters in all. Following the default setting, we pre-train the BERT-Large model ($L=24, H=1024$) with a sequence length of 128 on 8 Tesla V-100 GPUs. The experiments were implemented with the code from NVIDIA \footnote{\url{https://github.com/NVIDIA/DeepLearningExamples}} and mainly include two types of batch sizes: 8k and 32k, one of which represents the widely used setting for pre-training BERT and the other denotes the training scenario under large-batch training. The empirical results are presented in Figure \ref{fig:r1} and Figure \ref{fig:r2}. As illustrated in Figure \ref{fig:r1}, CAME achieves a significant improvement compared with Adam and Adafactor.  To be specific, CAME ($66.5\%$) increases validation accuracy at with an increment $3.4\%$ in comparison to Adafactor ($63.1\%$) using same number of training steps (20k). Apart from Adafactor, our proposed CAME achieves better performance than Adam in the pre-training of BERT-Large model with a huge reduction of memory cost.

% Generally, large-batch training leads to a wide gap in generalization which is much more challenging.
To evaluate the performance of our proposed CAME for large-batch training, we scale the batch size for BERT-Large training to 32,768 on Wikipedia dataset. As illustrated in Figure \ref{fig:r2}, CAME consistently reaches a more remarkable improvement compared with Adafactor. We notice that the accuracy of CAME on BERT-Large pre-training is $68.0\%$, which is highly over the original Adafactor ($61.9\%$) with same number of training steps. In addition, CAME reaches comparable accuracy with only half the training steps required for Adafactor. With batch size getting larger from 8k to 32k, CAME brings more enhancements to the training of BERT-Large in comparison with Adam and Adafactor. Compared with LAMB in large-batch training, CAME saves a high-level of memory footprint with slight training performance degradation.

\textbf{Memory Usage Comparison} \quad We set batch size to 1 to measure the memory usage of each optimizer more efficiently. As shown in Table \ref{tab:mc},  the two optimizers (Adam and LAMB) frequently employed for training large language models consume the highest amount of memory usage. Meanwhile, our proposed CAME optimizer exhibits a reduced memory footprint over the existing SM3 memory-efficient optimizer. As a consequence of our
confidence-guided strategy in CAME, there is no doubt that CAME will introduce an increased memory footprint in comparison with Adafactor. However, the extra memory footprint incurred of CAME is almost negligible ($1\%$) with a substantial performance improvement.

\begin{figure}[t]
    \centering
    \includegraphics[width=0.5\textwidth]{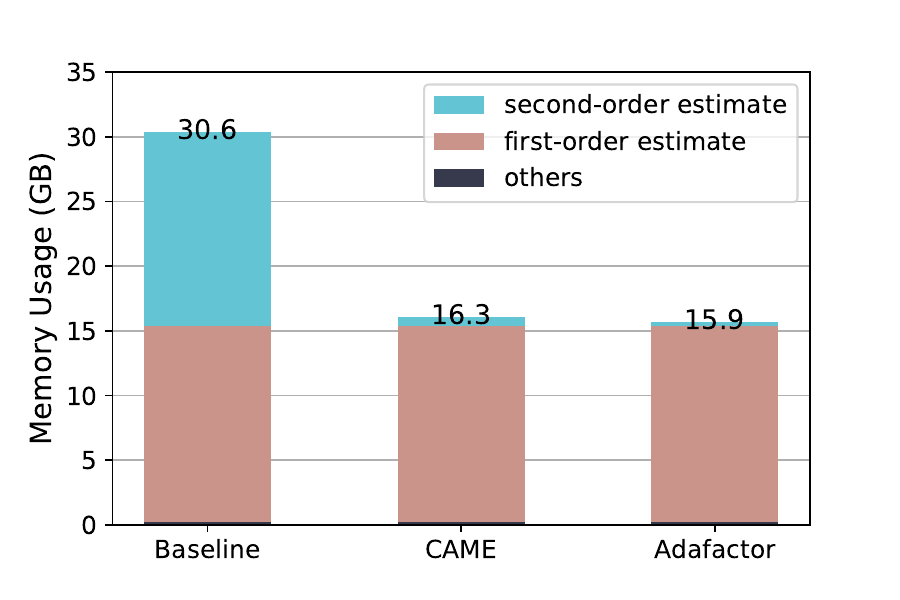}
    \caption{The memory reduction about optimizer states of CAME when training BERT-4B using PyTorch.
    % Trajectories of SGD, Adam and AdaBelief. AdaBelief reaches optimal point (marked as
% orange cross in 2D plots) the fastest in all cases. We refer readers to video examples.
}
    \label{fig:mc}
\end{figure}

For further demonstration of the memory saving effect of CAME, we expand BERT model to BERT-4B with 4 billion weights using the scaling method of GPT-3 \cite{10.5555/3495724.3495883}. We set the mini-batch size to 64 and the accumulation steps to 16 in this experiment. In Figure \ref{fig:mc}, we train BERT-4B with three different optimizers using PyTorch framework. As a result, CAME can save 47\% memory footprint about optimizer states compared with Baseline (Adam) when the weights number of a model get to 4 billion.

\noindent
\begin{table}[ht]
\centering
\caption{Quantitative memory usage per GPU (GB) comparison in the pre-training of BERT-Large model.}
\label{tab:mc}
    \begin{tabular}{ c  c }
      \toprule
      Optimizer  & Memory Cost (GB) \\
      \midrule
      Adam  & 8.24  \\
      LAMB  & 8.23 \\
      Adafactor  & 7.00  \\
      SM3  & 7.44  \\
      \midrule
      CAME  & 7.07  \\
      \bottomrule
    \end{tabular}
\end{table}

\begin{table*}[ht]

\centering
\caption{Results of fine-tuning performance on MNLI-m, SST-2, MRPC and two SQuAD datasets. The F1 and EM for SQuAD v1.1 dataset are firstly averaged, and the average of all results across five datasets is further calculated.}
\label{tab:ds}
    \begin{tabular}{ c c c c c c c }
      \toprule
      \textbf{Model} & \textbf{MNLI-m} & \textbf{SST-2} & \textbf{MRPC} & \textbf{SQuAD v1.1} & \textbf{SQuAD v2.0} & \textbf{Average}\\
      \textbf{} & (Acc) & (Acc) & (Acc) & (F1/EM) & (F1) & - \\
      \midrule
      Baseline  & 84.3 & 92.8 & 88.9 & 88.5/80.8 & 76.3 & 85.4 \\
      \midrule
      CAME (batch size = 8k) & 84.8  & 92.8 & 89.9 & 88.8/81.8 & 77.9 & 86.1 (\textbf{\textcolor[rgb]{0,0.5,0}{+0.7}}) \\
      CAME (batch size = 32k) & 84.5 & 92.9 & 89.8 & 88.5/81.2 & 77.4 &  85.9 (\textbf{\textcolor[rgb]{0,0.5,0}{+0.5}})\\
      \bottomrule
    \end{tabular}
\end{table*}

\subsection{Downstream Tasks}
We select a representative set of downstream tasks to further demonstrate the performance of BERT models pre-trained by our proposed CAME. In this part we adopt BERT-Base model for the fine-tuning task and follow the originally published
BERT-Base results in \cite{devlin-etal-2019-bert} and \cite{liu2019roberta} as the main baseline. The learning rate is tuned on the dev set for each setting and each task is fine-tuned for three epochs.
% In general, the empirical results show that the model trained with CAME obtains similar accuracy to the baseline.

We compare the end-task performance of BERT-Base with the baseline on typical downstream tasks and the empirical results are presented in Table \ref{tab:ds}. The experimental results demonstrate the efficiency of our proposed CAME optimizer by showing that BERT-Base model trained with CAME on two batch sizes both achieve comparable performance to the baseline with less memory cost. In particular, we observe that BERT-Base model trained with large batch (32k) presents no performance degradation and even attains higher evaluation metrics scores on some downstream tasks. 
Specifically, the BERT-Base model trained on CAME improves on average by 0.5 across five metrics compared to the baseline, 
proving the feasibility of CAME for the large-batch training task. 

\subsection{GPT-2 Training}
\label{sec:gpt}

In addition to BERT pre-training task, we perform CAME-based training task on another typical large language model, GPT-2. Using the original structure of GPT-2 \cite{gpt2}, we specifically adopt GPT-medium ($L=24, H=1024$) with 345M parameters in our experiment. This implementation is based on the code provided by Megatron\footnote{\url{https://github.com/NVIDIA/Megatron-LM}}. Identically, we take English Wikipedia as the training dataset for this section. Unlike the pre-training of BERT in Section \ref{sec:bert}, we only concentrate on standard training batch size (128) for GPT-2 pre-training. 

\begin{figure}[t]
    \centering
    \includegraphics[width=0.5\textwidth]{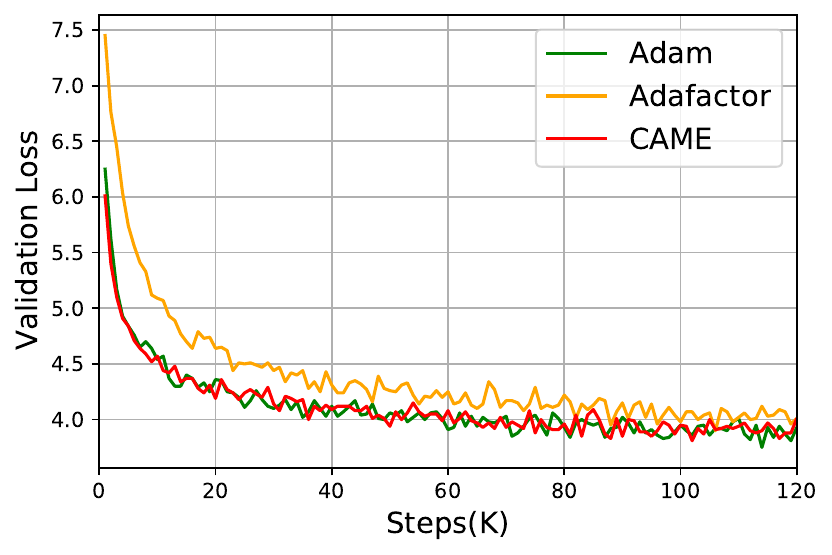}
    \caption{Validation loss of GPT-2 language model. CAME demonstrates similar optimization performance to Adam.}
    \label{fig:gpt1}
\end{figure}

\begin{figure}[t]
    \centering
    \includegraphics[width=0.5\textwidth]{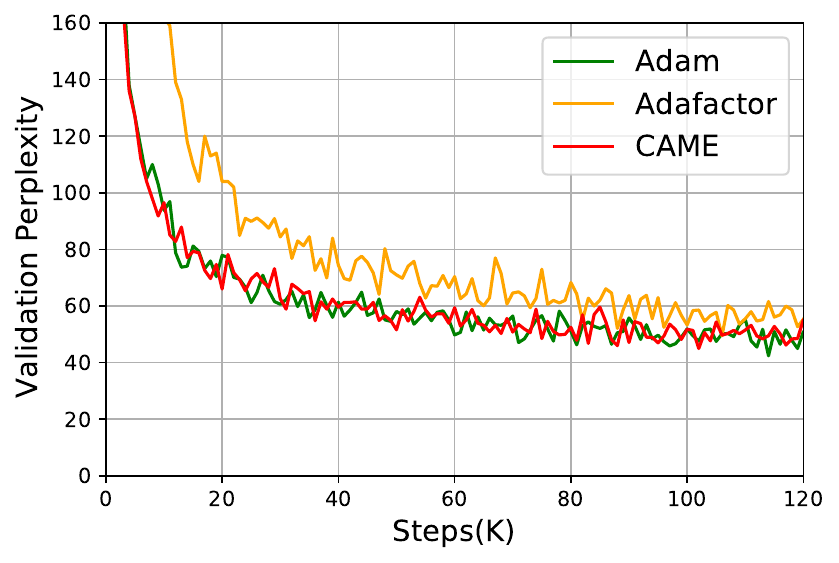}
    \caption{Validation perplexity of GPT-2 language model. CAME demonstrates comparable convergence speed with Adam.}
    \label{fig:gpt2}
\end{figure}

The empirical results of validation loss are shown in Figure \ref{fig:gpt1}. We are able to find that CAME achieves similar convergence and final accuracy compared to Adam, which reveals an impressive improvement over the performance of Adafactor with comparable training steps.
% Specifically, AdaRes achieves remarkably higher validation accuracy (80.7\%, 83.3\%, 84.4\%) compared with SAM-5 (80.4\%, 82.5\%, 83.8\%) on ResNet-18, ResNet-50 and WRN-28-10 respectively. 
Moreover, as indicated in Figure \ref{fig:gpt2}, the validation perplexity of CAME presents the same convergence performance as Adam but faster convergence speed than Adafactor, which clearly supports the validity of CAME that has fast convergence as in traditional adaptive methods and low memory usage as in existing memory-efficient methods. For instance, the converged validation perplexity of CAME and Adafactor is 50.1 and 56.9 respectively, which yields a considerable improvement of 12.0\%.

\subsection{T5 Training}
\label{sec:t5}
\begin{figure}[t]
    \centering
    \includegraphics[width=0.5\textwidth]{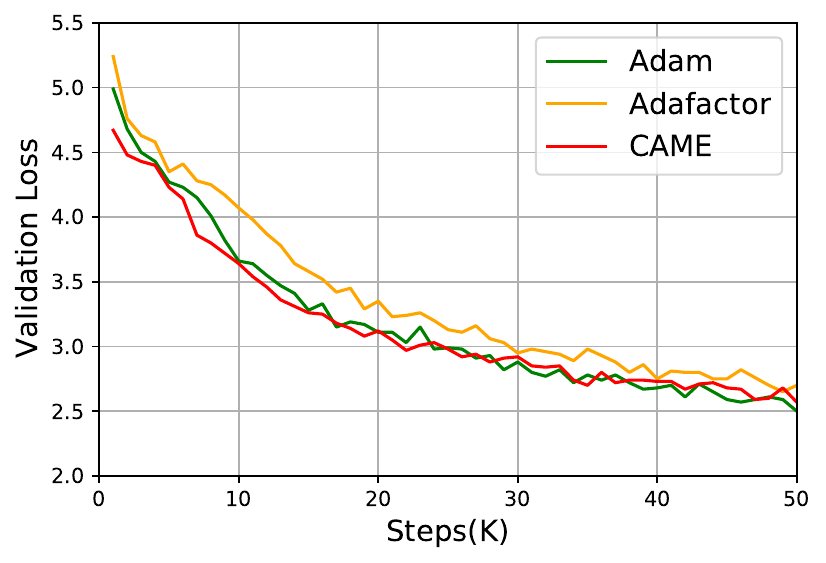}
    \caption{Validation loss of T5 language model. CAME exhibits similar convergence rates to Adam.}
    \label{fig:t5_1}
\end{figure}

\begin{figure}[t]
    \centering
    \includegraphics[width=0.5\textwidth]{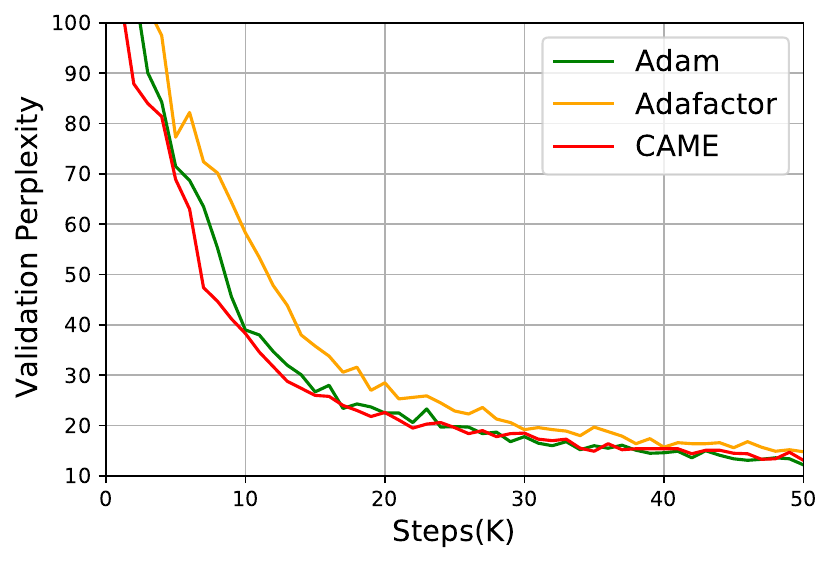}
    \caption{Validation perplexity of T5 language model. CAME demonstrates similar convergence speed to Adam.}
    \label{fig:t5_2}
\end{figure}

Finally, we report empirical results from a different large language model training task: Text-to-Text Transfer Transformer, T5. Concretely, we follow the architecture of T5 \cite{T5} and choose T5-Base ($L=24, H=1024$) with 220M parameters for the experiment. All of our implementations are also based on the code provided by Megatron. Similarly, we consider Wikipedia with 2.5B words as the training dataset in this part. As with the training of GPT-2 in Section \ref{sec:gpt}, we only concentrate on standard training batch size (128) for T5. 

The comparison of CAME with Adafactor and Adam is conducted in the same manner as Section \ref{sec:gpt}, and corresponding results of validation loss and validation perplexity are illustrated in Figure \ref{fig:t5_1} and Figure \ref{fig:t5_2} seperately. Note that CAME consistently obtains comparable convergence performance for validation loss and validation perplexity on par with Adam, while reducing similar memory usage as Adafactor. 
% We are able to find that CAME achieves similar convergence and final accuracy compared to Adam, which reveals an impressive improvement over the performance of Adafactor with the comparable training steps.
% Specifically, AdaRes achieves remarkably higher validation accuracy (80.7\%, 83.3\%, 84.4\%) compared with SAM-5 (80.4\%, 82.5\%, 83.8\%) on ResNet-18, ResNet-50 and WRN-28-10 respectively. 
% Moreover, as indicated in Figure \ref{fig:t5_2}, the validation loss of CAME presents the same convergence performance of three optimizers, which clearly supports the validity of CAME that has fast convergence as in traditional adaptive methods and low memory usage as in exsiting memory efficient methods.
% According to Figure \ref{fig:t5_2}, the validation perplexity of CAME is equal to the convergence performance of Adam, thus demonstrating its validity in the optimization process. In comparison to traditional adaptive methods, CAME achieves fast convergence and requires less memory footprint than existing memory efficient algorithms.

\section{Conclusion}
In this paper we propose a novel memory-efficient optimizer called CAME, which supports adaptive confidence-based updating guided by the residual between predicted update and generated update. CAME achieves a considerable improvement compared to existing memory-efficient optimizers in the training of large language models, with an ignorable extra memory footprint. Moreover, CAME shows comparable convergence to Adam and LAMB with huge memory reduction. In particular, CAME has proven effective for large-batch training, which serves as an advantageous extension to memory-efficient optimizers. We hope our work will provide insight into memory reduction of optimizers in future exploration.

\section{Limitations}
Despite the success of our CAME optimizer in training large language models with memory efficiency, there are still some limitations that need to be addressed in the future.

Our proposed memory-efficient optimizer introduces additional computation costs for the non-negative matrix factorization of the instability matrix in comparison with Adafactor. We observe, however, that the training time of CAME increases only slightly in our experiments. 
% Beyond that, CAME exhibits minor performance degradation in large-batch training of the BERT-Large model versus LAMB, which allows for further improvement in the future. 
Beyond that, CAME exhibits minor performance degradation in large-batch training of the BERT-Large model versus LAMB, which allows for further improvement in the future. Meanwhile, it is possible to conduct further experiments on other models in other fields, such as Computer Vision and Reinforcement Learning, thereby exploring the effectiveness of CAME training under more application scenarios. As a final point, it would be much more helpful to provide an in-depth theoretical analysis of CAME to improve comprehensiveness of the paper.

\section*{Acknowledgements}
Yang You's research group is being sponsored by NUS startup grant (Presidential Young Professorship), Singapore MOE Tier-1 grant, ByteDance grant, ARCTIC grant, SMI grant and Alibaba grant. We also thank Huawei Noah's Ark Lab for providing the necessary computing resources and support for datasets.

% This document has been adapted
% by Steven Bethard, Ryan Cotterell and Rui Yan
% from the instructions for earlier ACL and NAACL proceedings, including those for 
% ACL 2019 by Douwe Kiela and Ivan Vuli\'{c},
% NAACL 2019 by Stephanie Lukin and Alla Roskovskaya, 
% ACL 2018 by Shay Cohen, Kevin Gimpel, and Wei Lu, 
% NAACL 2018 by Margaret Mitchell and Stephanie Lukin,
% Bib\TeX{} suggestions for (NA)ACL 2017/2018 from Jason Eisner,
% ACL 2017 by Dan Gildea and Min-Yen Kan, 
% NAACL 2017 by Margaret Mitchell, 
% ACL 2012 by Maggie Li and Michael White, 
% ACL 2010 by Jing-Shin Chang and Philipp Koehn, 
% ACL 2008 by Johanna D. Moore, Simone Teufel, James Allan, and Sadaoki Furui, 
% ACL 2005 by Hwee Tou Ng and Kemal Oflazer, 
% ACL 2002 by Eugene Charniak and Dekang Lin, 
% and earlier ACL and EACL formats written by several people, including
% John Chen, Henry S. Thompson and Donald Walker.
% Additional elements were taken from the formatting instructions of the \emph{International Joint Conference on Artificial Intelligence} and the \emph{Conference on Computer Vision and Pattern Recognition}.

\bibliography{anthology,custom}
\bibliographystyle{acl_natbib}

\end{document}